\documentclass[journal=ancham,manuscript=article]{achemso}

\usepackage[T1]{fontenc}
\usepackage{booktabs}
\usepackage{graphicx}
\usepackage{float}
\usepackage{multirow}
\usepackage{amsmath}
\usepackage{amssymb}
\usepackage{hyperref}
\usepackage{tabularx}
\usepackage{xcolor}
\hypersetup{hidelinks}
\let\EnableSupplementSectionNumbers\SectionNumbersOn
\author{Yinkai Wang}
\affiliation{Department of Computer Science, Tufts University, Medford, MA, 02155, USA}

\author{Yan Zhou Chen}
\affiliation{Department of Computer Science, Tufts University, Medford, MA, 02155, USA}

\author{Soha Hassoun}
\altaffiliation{Corresponding author}
\email{soha.hassoun@tufts.edu}
\affiliation{Department of Computer Science, Tufts University, Medford, MA, 02155, USA}
\alsoaffiliation{Department of Chemical and Biological Engineering, Tufts University, Medford, MA, 02155, USA}

\title[MS-AGG] {MS-AGG: Learning to Aggregate Tandem Mass Spectrometry Annotation Scores and Features for Improved Molecular Retrieval}

\begin{document}

\begin{abstract}

Identifying molecular structures from tandem mass spectra remains a central challenge in untargeted metabolomics. Computational annotation methods use different representations and sources of information to rank candidate molecular structures, creating opportunities to improve retrieval by combining their predictions. However, existing aggregation approaches typically rely on fixed rules for combining annotation scores and do not fully exploit additional information about candidate structures. Learned aggregation offers a way to optimize how these signals are combined for molecular retrieval.
We introduce MS-AGG (Mass-Spectrometry Annotation Aggregation), a framework that learns to combine annotation-model scores with candidate-level features derived from predicted molecular fingerprints and spectrum--structure reference neighborhoods.
The aggregation function is trained to optimize relative candidate rankings within each query while keeping the constituent annotation models fixed.  We evaluate MS-AGG across five candidate-pool settings on MassSpecGym and Spectraverse, covering mass-based, formula-restricted, and generated same-formula decoy pools. 
MS-AGG rankers outperform the strongest individual annotation method, all fixed-rule aggregation baselines, and the learned pointwise baseline in every setting. For example, on MassSpecGym formula pools, MS-AGG-LambdaMART achieves $46.98 \pm 0.21\%$ Hit@1, compared with $35.94\%$ for the strongest constituent model and $40.38\%$ for pointwise logistic regression. Further, feature ablations demonstrate the additional value of derived candidate-level information. These results demonstrate that molecular retrieval benefits from both learned aggregation that explicitly optimizes relative candidate rankings and the incorporation of additional predictive information not captured by individual annotation scores.
\end{abstract}

\section*{Introduction}
Untargeted metabolomics aims to characterize small molecules in complex biological samples. 
However, identifying the molecular structures underlying observed tandem mass spectra (MS/MS) remains a major challenge in translating metabolomic measurements into biological insights \cite{Theodoridis2026MetaboliteIdentification,Houriet2025AnnotationBottlenecks}. Reference libraries cover only part of the chemical diversity encountered in samples, limiting the applicability of direct spectral-library matching\cite{Hoffmann2022COSMIC}. Further, differences in instrumentation and acquisition conditions can reduce the reliability of spectral-library matching\cite{bittremieux2022critical}, while highly similar MS/MS spectra from distinct molecular structures can lead to ambiguous or incorrect matches.\cite{Hoffmann2022COSMIC}. Consequently, analyzing query spectra without direct or near spectral-library matches often relies on computational molecular retrieval, in which candidate structures from chemical databases are ranked according to their compatibility with the observed MS/MS spectrum.

Computational methods address the molecular retrieval problem through fingerprint prediction\cite{Heinonen2012FingerID,Duhrkop2015CSIFingerID,Goldman2023MIST}, including predictors built on pretrained spectral encoders\cite{Bushuiev2026DreaMS}; in silico fragmentation and spectrum prediction\cite{Ruttkies2016MetFrag,Wang2021CFMID4,Li2024ESP,Goldman2024ICEBERG,Young2024FraGNNet,Nowatzky2025FIORA,Wang2026GLACIER}; and cross-modal spectrum--structure alignment\cite{Kalia2025JESTR,Chen2026FLARE,Chen2025MVP,Krzakala2026MSAlign}. Candidate molecules can be retrieved from large molecular databases such as PubChem\citep{Kim2025pubchem}, and potentially restricted by chemical formula \cite{Duhrkop2019SIRIUS,Xing2023BUDDY} or to biologically relevant molecules \citep{hassanpour2020biological}. Although these restrictions reduce the number of candidate structures, distinguishing among the remaining candidates remains challenging.

Different computational annotation methods may rank ambiguous candidate structures differently: a query that is difficult for one annotation method may be easier for another. FLARE\cite{Chen2026FLARE}, for example, outperforms JESTR\cite{Kalia2025JESTR} on average, yet JESTR ranks the target molecule substantially higher for some queries\cite{Chen2026FLARE}. This disagreement suggests that the two tools capture complementary information. MetaboAnnotate  \cite{Chen2026MetaboAnnotate} exploits this complementarity by combining rankings from SIRIUS\cite{Duhrkop2019SIRIUS}, FLARE, and JESTR using  reciprocal rank fusion, weighted averaging, and Borda count. Evaluations show gains in top-$k$ retrieval on the CASMI datasets, and that requiring agreement among tools reduces false discovery rates at the cost of annotation coverage. Similarly, combining FraGNNet\cite{Young2024FraGNNet}, ICEBERG\cite{Goldman2024ICEBERG}, FIORA\cite{Nowatzky2025FIORA}, MIST\cite{Goldman2023MIST}, and FLARE using rank products\cite{breitling2004rank} improves candidate ranking over any individual tool \cite{gupta2026confronting}. 
These studies demonstrate that aggregating predictions from multiple annotation methods can improve molecular retrieval when using fixed aggregation rules.

Unlike predefined aggregation rules, learned aggregation can adaptively weight annotation scores based on their informativeness and redundancy. Early work explored this approach by learning linear combinations of candidate scores from variants of the input--output kernel regression framework\citep{Brouard2019IOKRfusion}. More recently, MSAlign introduced a validation-optimized convex combination of multiple spectral–molecular alignment scores and scores derived from the mass difference between each candidate molecule and the observed precursor\cite{Krzakala2026MSAlign}. These approaches demonstrate the potential of learning how to combine scores. Beyond learning how to combine scores, the choice of the aggregation objective is also important, as optimizing surrogate objectives such as fingerprint similarity does not necessarily improve molecular retrieval\cite{DeWaele2026Optimizing}. 
Learning-to-rank methods provide a natural framework for optimizing molecular retrieval directly. Widely used in information retrieval to rank documents by their relevance to a query \citep{liu2009learning}, these methods can similarly rank candidate molecular structures for a query spectrum by learning how to combine scores from multiple annotation methods.

We introduce in this paper MS-AGG (\emph{Mass-Spectrometry Annotation Aggregation}), a learning-to-rank framework for molecular retrieval for tandem mass spectra. Rather than relying on predefined aggregation rules,  MS-AGG learns to combine scores from multiple annotation methods to rank the candidates by optimizing relative candidate rankings within each query.  Because annotation scores may not capture all the information useful for distinguishing candidate structures, the framework also incorporates "derived features" that quantify fingerprint agreement and relationships between candidate molecules, query spectra, and reference spectrum-molecule pairs retrieved from spectral libraries. The aggregation models are trained on labeled validation spectra using predicted scores from the constituent annotation sources. MS-AGG therefore learns how to integrate information without modifying or retraining the underlying models.

We evaluate MS-AGG on MassSpecGym\cite{Bushuiev2024MSGym} and Spectraverse\cite{gupta2026comprehensive} across five candidate-pool settings covering mass-based, formula-restricted, and generated same-formula decoy pools. 
We compare MS-AGG with individual annotation methods, fixed aggregation rules, and a learned pointwise baseline to assess the benefits of ranking-based learned aggregation.
Through ablation studies, we assess the contributions of derived features and individual methods. Finally, query-level analyses characterize the benefits and limitations of learned aggregation by examining its ability to recover target structures not ranked first by any individual feature and assessing its performance relative to a retrospective feature-selection oracle.

\section*{Methods}

\subsection*{MS-AGG Framework}
\begin{figure}[p]
\centering
\includegraphics[width=\linewidth]{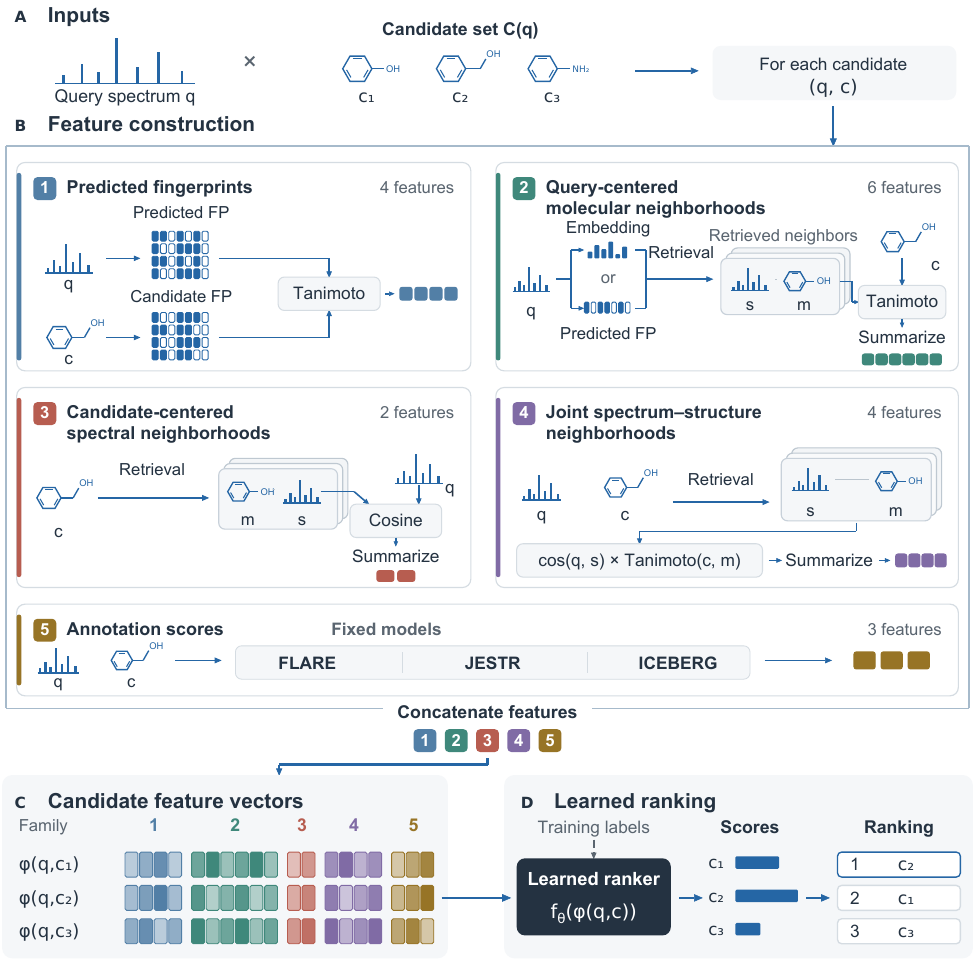}
\caption{MS-AGG overview. (A) A query spectrum $q$ and candidate set $C(q)$ define spectrum--candidate pairs $(q,c)$. (B)  Five feature families are constructed for each pair: four derived-feature families and annotation-model scores. Spectral-library retrieval based on embedding, fingerprint, or spectral similarity results in  reference spectrum-molecule  pairs.  (C) Features are concatenated into a candidate-level vector $\phi(q,c)$. (D) A learned aggregator produces a ranking of candidates $c_1, c_2, ... $ at inference. }
\label{fig:msagg-representation}
\end{figure}

For a query spectrum, MS-AGG ranks a  set of candidate molecular structures by aggregating scores from multiple annotation methods and additional derived features (Figure~\ref{fig:msagg-representation}). For each spectrum–candidate pair, the framework constructs a feature vector corresponding to five feature families.  A ranking model, trained on labeled validation spectra while keeping the constituent annotation models fixed, learns to combine these features by optimizing relative candidate rankings within each query. For a new query spectrum, each of the tools and family comparisons is performed. The trained aggregation model then assigns a score to each candidate, and the candidates are ranked by these scores.

Each feature family is designed to capture complementary sources of information for assessing the agreement between a query spectrum and a candidate molecule. Predicted-fingerprint features compare  molecular fingerprints predicted from the query spectrum with the corresponding fingerprints calculated from each candidate structure. Query-centered molecular neighborhood features first retrieve top reference molecules from spectral libraries using  spectral similarity of the query and its predicted-fingerprint similarity, then measure how structurally similar those molecules are to the candidate. Candidate-centered spectral neighborhood features retrieve from spectral libraries  reference spectra whose associated molecules are structurally similar to each candidate, then measure the spectral similarity of these retrieved spectra to the query spectrum. Joint spectrum–structure neighborhood features  build on the query-centered and candidate-centered retrieval approaches, but score the spectral similarity to the query and structural similarity to the candidate for each retrieved spectrum-molecular reference pair. The last three families do not require the query molecule to be present in the spectral library. Instead, they exploit spectral and structural relationships to known reference molecules in spectral libraries to provide additional information for ranking candidate structures. The final family contains scores from existing annotation methods, FLARE, JESTR, and, where available, ICEBERG, that evaluate how well each candidate structure matches the query spectrum. Within each family, multiple features (Table~\ref{tab:features}) summarize different aspects of spectrum–structure agreement.

\begin{table}[!htbp]
\caption{Candidate-level features used by MS-AGG.}
\label{tab:features}
\centering
\footnotesize
\begin{tabular}{>{\raggedright\arraybackslash}p{0.27\textwidth}c>{\raggedright\arraybackslash}p{0.56\textwidth}}
\toprule
Family & Count & Feature names \\
\midrule
1 {Predicted fingerprints} & 4 & \texttt{s\_r3}, \texttt{s\_r2}, \texttt{s\_av}, \texttt{s\_ap} (Morgan r=3, Morgan r=2, Avalon, Atom-Pair Tanimoto to the predicted fingerprint) \\
2 Query-centered molecular neighborhoods & 6 & \texttt{nn\_max\_K10}, \texttt{nn\_max\_K30}, \texttt{nn\_max\_K100}, \texttt{nn\_mean\_K100} (Tanimoto similarities to spectral neighbors); \texttt{pn\_msn}, \texttt{pn\_msn\_spec} (maximum Morgan Tanimoto to neighbors retrieved using predicted fingerprints and spectral embeddings, respectively) \\
3 Candidate-centered spectral neighborhoods & 2 & \texttt{s\_anchor\_max}, \texttt{s\_anchor\_posw} (query--reference spectral cosine over structural neighbors: max, and fingerprint-similarity-weighted mean of cosines clipped at zero) \\
4 Joint spectrum--structure neighborhoods & 4 & \texttt{joint\_spec\_max}, \texttt{joint\_mol\_max}, \texttt{joint\_mol\_wmean}, \texttt{joint\_mol\_top3} \\
5 Annotation scores & 2 or 3 & \texttt{s\_flare}, \texttt{s\_iceberg}, \texttt{s\_jestr} \\
\bottomrule
\end{tabular}
\begin{minipage}{0.98\linewidth}
\footnotesize Families 1--4 provide 16 derived features; family 5 supplies two or three annotation scores as MassSpecGym uses FLARE, ICEBERG, and JESTR (19 features) and Spectraverse uses FLARE and JESTR (18 features).
\end{minipage}
\end{table}

\subsection*{Problem Formulation}
Given a query MS/MS spectrum $q$, let ${C(q)=\{c_1,\ldots,c_{n_q}\}}$ denote a set of candidate molecular structures. MS-AGG ranks these candidates, which may be obtained using precursor mass, predicted or known molecular formula, or another retrieval strategy. For each spectrum–candidate pair $(q,c)$, MS-AGG constructs a feature vector $\phi(q,c)$. A learned ranking function $f_\theta$ assigns each candidate a score $s_\theta(q,c)=f_\theta(\phi(q,c))$, and candidates are ranked in descending order of their scores.

For training and retrieval evaluation, the target structure $m_q^*$  is known and included in $C(q)$. Each candidate receives a binary relevance label $y_{q,c}=1$ if its canonical SMILES matches that of $m_q^*$, and $y_{q,c}=0$ otherwise. The ranking function is trained to place the target structure above incorrect candidates. At inference, the target structure is unknown, and MS-AGG ranks the supplied candidates without requiring that the target be present.

\subsection*{Feature Construction}
\paragraph{Predicted molecular fingerprints.} A multi-fingerprint prediction head built on the pretrained DreaMS encoder\cite{Bushuiev2026DreaMS} maps the query spectrum to fingerprints.
We use four fingerprint types: Morgan r = 2 and r = 3, which describe local atomic environments at different radii\cite{Rogers2010ECFP}; Avalon\cite{gedeck2006qsar}, which encodes molecular fragments and structural patterns; and Atom-Pair\cite{carhart1985atom}, which captures relationships between pairs of atoms and their distances along the molecular graph. The  fingerprints capture different aspects of molecular structure, providing complementary measures of agreement between the query spectrum and each candidate.

\paragraph{Query-centered molecular neighborhoods.} 
Six features measure the structural similarity between each molecular candidate and reference molecules retrieved based on the query spectrum. The features differ in how reference molecules are retrieved and how their similarities to the candidate are summarized. In all cases, candidate–reference structural similarity is measured using Morgan fingerprint Tanimoto similarity.

The first four features use spectral similarity to retrieve reference molecules from a reference spectral library. Specifically, cosine similarity between DreaMS embeddings is used to identify the 100 nearest distinct reference molecules to the query spectrum. 
Three features report the maximum candidate–reference Tanimoto similarity among the top 10, 30, and 100 retrieved molecules, respectively. The fourth reports the mean similarity among the top 100. The remaining two features use different retrieval procedures. For \texttt{pn\_msn}, cosine similarity between the query’s \textit{predicted} Morgan $r=3$ fingerprint and reference fingerprints is used to retrieve 100 molecules from the reference library.  For \texttt{pn\_msn\_spec}, cosine similarity between DreaMS embeddings is used to retrieve 100 reference spectra. The two features report the maximum candidate–reference Tanimoto similarity among their respective retrieved references. In our evaluation, we ensured that the target molecule was not available from the reference library for retrieval.

\paragraph{Candidate-centered spectral neighborhoods.}
For each candidate $c$, we retrieve the 20 reference entries from the training set whose associated molecules have the highest Tanimoto similarity to $c$. Multiple spectra from the same molecule may be included. 

For each retrieved reference spectrum $s_i$ and its associated molecule $m_i$, 
we define spectral similarity $a_i=\cos(\mathrm{emb}(q),\mathrm{emb}(s_i))$ and structural weight $w_i=\mathrm{Tanimoto}(\mathrm{FP}(c),\mathrm{FP}(m_i))$. Embeddings are calculated using the DreaMS spectral encoder and the Tanimoto similarity on the fingerprint (FP) is calculated based on Morgan Fingerprint,  radius 2, 4096 bits.  The two features are as follows:
\[
\texttt{s\_anchor\_max}=\max_i a_i,\qquad
\texttt{s\_anchor\_posw}=\frac{\sum_i w_i\,\max(a_i,0)}{\sum_i w_i}.
\]
The first feature reports the highest spectral similarity among the retrieved references. The second reports a weighted average of spectral similarities, giving greater weight to references whose molecules are more structurally similar to the candidate. Negative spectral similarities are set to zero before averaging. If all structural weights are zero,
\texttt{s\_anchor\_posw} is set to zero.

\paragraph{Joint spectrum--structure neighborhoods.}  For each of the 100 query spectral neighbors used to construct the \texttt{nn\_max\_K100} feature, we multiply its spectral cosine similarity to the query by its molecule's Morgan Tanimoto similarity to the candidate. The largest product gives \texttt{joint\_spec\_max}.
We also retrieve the 20  training molecules that are most structurally similar to each candidate. For each such molecule, we retain the highest query similarity among its reference spectra and multiply it by the molecule's Tanimoto similarity to the candidate. We summarize these products by their maximum (\texttt{joint\_mol\_max}), the mean of the three largest values (\texttt{joint\_mol\_top3}), and a weighted mean using the candidate--reference Tanimoto similarities as weights (\texttt{joint\_mol\_wmean}).

\paragraph{Annotation scores.} FLARE,\cite{Chen2026FLARE} ICEBERG,\cite{Goldman2024ICEBERG} and JESTR\cite{Kalia2025JESTR} each contribute a candidate-level score, based on peak-to-substructure alignment, spectrum simulation, and contrastive spectrum--molecule embeddings, respectively. The models are trained on relevant benchmark's training set. 
For ICEBERG, we use the released ICEBERG 2.1 \texttt{msg\_all} models. 
Each candidate is scored by spectral entropy similarity\cite{Li2021SpectralEntropy} using the query adduct and observed collision energy, with a fixed 30~eV default when energy is missing.

\subsection*{MS-AGG Ranking Models}
We develop four MS-AGG variants that differ in their architectures and training objectives. All models use pairwise or listwise learning objectives that account for the relative rankings of candidates within each query. The linear pairwise ranker learns feature weights by comparing target-matching and non-matching candidates from the same query.
The multilayer perceptron (MLP) is a neural network that learns nonlinear relationships between candidate-level features to predict a ranking score for each candidate. The MLP is trained using a query-wise softmax cross-entropy loss, which converts candidate scores into a probability distribution over the candidates for each query and penalizes the model when it assigns low probability to the correct molecular structure. This encourages the model to assign higher scores to the correct structure relative to competing candidates.

LambdaMART~\cite{Burges2010LambdaMART,Ke2017LightGBM} is a gradient-boosted tree ensemble that learns to rank candidates through pairwise comparisons. These comparisons are weighted according to how much swapping the positions of two candidates would change normalized discounted cumulative gain (NDCG), a ranking metric that rewards placing relevant candidates near the top of the list. Consequently, LambdaMART emphasizes correcting ranking errors that affect the highest-ranked candidates. We implement LambdaMART using LightGBM, a gradient-boosting framework that provides the \texttt{lambdarank} objective.

YetiRank~\cite{Gulin2011YetiRank,Prokhorenkova2018CatBoost} is also a gradient-boosted tree ensemble that learns to rank candidates through pairwise comparisons. Unlike LambdaMART, which weights candidate pairs according to their effect on NDCG, YetiRank samples possible candidate orderings to estimate the importance of different pairwise comparisons. These weights guide the model toward correcting ranking errors that are more consequential for the overall ranking. We implement YetiRank using CatBoost, a gradient-boosting framework that provides the YetiRank objective in its default Classic mode.

All four models use the same candidate-level features and produce one score per candidate. The constituent annotation models remain fixed, and only the parameters of the aggregation models are learned. For each candidate setting, we fit all  models on the same 90\% of benchmark validation queries and select the model with the highest Hit@1 on the remaining 10\%. For validation, we use average candidate scores across four runs using different seeds for each of the  models. We evaluate the selected model on the independent test set. Implementation details are provided in Section~\ref{si:ranker-methods}.

\subsection*{Baseline Aggregation Models}
We evaluate two types of aggregation baselines:  pointwise learned aggregation and fixed-rule aggregation. These baselines allow us to assess the benefits of learning the aggregation function and of optimizing relative candidate rankings within each query.

For learned pointwise aggregation, we use logistic regression with the same input features and training queries as the MS-AGG rankers. The model learns feature weights by minimizing binary cross-entropy to distinguish target-matching from non-matching candidates. At inference, candidates within each query are ranked by their predicted scores. Unlike the MS-AGG rankers, pointwise logistic regression learns an aggregation function without explicitly optimizing relative candidate rankings during training.

For fixed-rule aggregation, we consider previously used methods \cite{gupta2026confronting,Chen2026MetaboAnnotate}, which  rely on candidate ranks rather than scores. Rank product multiplies each candidate's ranks across the input features, with lower products indicating better aggregate rankings. Reciprocal rank fusion sums the inverse of each candidate's rank plus a fixed constant across input features, with higher sums indicating better aggregate rankings. Borda count aggregates rankings by assigning points to candidates according to their positions in each ranking. 

To evaluate an alternative that retains information about score magnitudes, we use z-mean aggregation, which averages standardized candidate-level features with equal weights. We consider two standardization approaches. Within-pool z-mean standardizes each feature relative to the scores of competing candidates for the same query, emphasizing how strongly a candidate stands out within its pool. In contrast, global z-mean uses statistics estimated across validation candidates, placing feature scores on a common scale across queries. All such baseline approaches can produce different rankings because they change the relative contributions of individual features to the aggregated score. Further implementation details are provided in Section~\ref{si:fusion}.

\subsection*{Evaluation Metrics}
We report Hit@$k$ for $k\in\{1,5,20\}$, and MCES@1. Hit@$k$ is the fraction of queries for which the target structure occurs among the first $k$ candidates. The primary retrieval results use 2D-connectivity identity, determined by the first block of the InChIKey. MCES@1 measures the structural distance between the top-ranked candidate and the target structure, with smaller values indicating greater structural similarity. %

\section*{Results}

\subsection*{Benchmark Datasets and Candidate Sets}
We evaluate MS-AGG across five candidate-pool settings (Table~\ref{tab:datasets})  
on MassSpecGym and Spectraverse: database-derived mass- and formula-based pools and generated same-formula decoys. The decoy pools complement database-derived candidates, whose distribution can differ from that of molecules with experimental spectra\cite{gupta2026confronting,Liu2026MassSpecGymWild}.

MassSpecGym provides mass- and formula-based pools for 17{,}556 test spectra. We construct same-formula decoy pools using a character-level chemical-language model (CLM)\cite{Skinnider2021CLM} trained on MassSpecGym training structures, supplementing our generated candidates with released Spectraverse CLM candidates\cite{gupta2026confronting}. Formula matching and 2D deduplication yield decoy pools for 16{,}812 test spectra. In the mass-based setting, FLARE and the five formula-dependent derived features receive formulas predicted by the MIST-CF checkpoint\cite{Goldman2024MISTCF} released with MassSpecGym v1.5\cite{Liu2026MassSpecGymWild}. The predicted formula matches the target formula for 9,495 of the 17,556 test queries (54.1\%). All three settings use at most 256 candidates per spectrum.  

Spectraverse provides two formula-restricted settings: PubChem-derived pools with up to 32{,}280 candidates per spectrum, and released CLM-generated decoy pools\cite{gupta2026comprehensive,gupta2026confronting}. We evaluate spectra whose target is present in the pool: 46{,}042 test spectra for PubChem and 45{,}132 for CLM decoys. 

The reference libraries contain benchmark training spectra and, for MassSpecGym, additional spectra from MS$^n$Lib\cite{brungs2025msnlib}. Both benchmarks use 16 derived features and the benchmark-specific FLARE and JESTR scores. ICEBERG is excluded from Spectraverse benchmark evaluations because its training structures match $36.6\,\%$ of the Spectraverse target structures.
For MassSpecGym formula pools, FLARE and JESTR results were taken from their respective publications; the published JESTR evaluation covers 17,368 spectra rather than the full 17,556. Additional  dataset analyses  are provided in Section \ref{sec:suppData}.

\subsection*{MS-AGG Outperforms Individual Annotation Tools}

All four MS-AGG rankers outperform the strongest individual annotation tool in Hit@1 across all five candidate-pool settings  (Table~\ref{tab:retrieval}). For example, on MassSpecGym formula-restricted pools, the linear pairwise ranker achieves $46.25 \pm 0.02\%$ Hit@1, compared with $35.94\%$ for ICEBERG. On Spectraverse PubChem pools, the MLP achieves $41.73 \pm 0.42\%$ Hit@1, compared with $18.74\%$ for FLARE.

No single aggregation architecture performs best across all candidate-pool settings. While the MLP achieves the highest Hit@1 on both Spectraverse candidate pools, the linear pairwise ranker achieves the highest Hit@1 ($49.88 \pm 0.06\%$) and the lowest MCES@1 on MassSpecGym CLM decoys. The strong performance of the linear pairwise ranker indicates that nonlinear aggregation is not always necessary to achieve competitive retrieval performance. These differences motivate selecting the aggregation model separately for each candidate-pool setting.

The validation-selected models, indicated by a dagger in Table~\ref{tab:retrieval}, consistently outperform the strongest individual annotation tool across all five settings, demonstrating that validation-based model selection yields reliable improvements across diverse candidate pools. Additional analyses, including statistical comparisons, performance across ionization modes and adducts, and cross-dataset feature transfer, are provided in Section~\ref{sec:suppResults}.

\begin{table}[!htbp]
\caption{Learned Aggregation Methods Outperform Individual Annotation Tools.}
\label{tab:retrieval}
\centering
\scriptsize
\setlength{\tabcolsep}{5pt}
\renewcommand{\arraystretch}{0.72}

\begin{tabular}{lrrrr}
\toprule
Method & Hit@1 (\%) $\uparrow$ & Hit@5 (\%) $\uparrow$ & Hit@20 (\%) $\uparrow$ & MCES@1 $\downarrow$ \\
\midrule
\multicolumn{5}{l}{\textbf{MassSpecGym, mass-based pools} ($n=17{,}556$)} \\

FLARE\cite{Chen2026FLARE} & 24.72 & 46.72 & 65.78 & 18.50 \\
ICEBERG\cite{Goldman2024ICEBERG} & 46.15 & 61.53 & 72.25 & 15.79 \\
JESTR\cite{Kalia2025JESTR} & 9.14 & 24.34 & 44.29 & 23.00 \\
MS-AGG & & & & \\
\quad Linear pairwise & 53.31 $\pm$ 0.02 & 75.05 $\pm$ 0.03 & 87.97 $\pm$ 0.01 & 9.93 $\pm$ 0.01 \\
\quad MLP & \textbf{54.66 $\pm$ 0.61} & 75.77 $\pm$ 0.27 & 88.44 $\pm$ 0.23 & 9.46 $\pm$ 0.07 \\
\quad YetiRank & 54.26 $\pm$ 0.11 & \textbf{76.05 $\pm$ 0.11} & \textbf{89.09 $\pm$ 0.12} & \textbf{8.85 $\pm$ 0.03} \\
\quad LambdaMART$^{\dagger}$ & 52.45 $\pm$ 0.34 & 74.88 $\pm$ 0.23 & 89.03 $\pm$ 0.04 & 9.16 $\pm$ 0.15 \\
\midrule
\multicolumn{5}{l}{\textbf{MassSpecGym, formula pools} ($n=17{,}556$)} \\

FLARE & 22.66 & 50.00 & 75.15 & 9.00 \\
ICEBERG & 35.94 & 59.60 & 78.29 & 6.48 \\
JESTR & 11.85 & 32.95 & 61.46 & 11.68 \\
MS-AGG & & & & \\
\quad Linear pairwise & 46.25 $\pm$ 0.02 & 72.57 $\pm$ 0.05 & \textbf{89.24 $\pm$ 0.01} & 5.11 $\pm$ 0.00 \\
\quad MLP & 46.31 $\pm$ 0.90 & 72.89 $\pm$ 0.58 & 88.50 $\pm$ 0.52 & 4.99 $\pm$ 0.14 \\
\quad YetiRank & \textbf{47.33 $\pm$ 0.07} & 72.82 $\pm$ 0.28 & 88.38 $\pm$ 0.06 & 4.96 $\pm$ 0.01 \\
\quad LambdaMART$^{\dagger}$ & 46.98 $\pm$ 0.21 & \textbf{73.07 $\pm$ 0.19} & 88.21 $\pm$ 0.16 & \textbf{4.93 $\pm$ 0.03} \\
\midrule
\multicolumn{5}{l}{\textbf{MassSpecGym, CLM decoys} ($n=16{,}812$)} \\
FLARE & 20.01 & 44.63 & 66.41 & 11.08 \\
ICEBERG & 42.68 & 62.37 & 76.75 & 7.37 \\
JESTR & 9.59 & 24.79 & 45.78 & 13.54 \\
MS-AGG & & & & \\
\quad Linear pairwise & \textbf{49.88 $\pm$ 0.06} & \textbf{69.37 $\pm$ 0.01} & 81.53 $\pm$ 0.02 & \textbf{6.46 $\pm$ 0.01} \\
\quad MLP & 49.05 $\pm$ 0.50 & 69.31 $\pm$ 0.38 & \textbf{82.31 $\pm$ 0.47} & 6.59 $\pm$ 0.06 \\
\quad YetiRank$^{\dagger}$ & 46.00 $\pm$ 0.55 & 67.38 $\pm$ 0.45 & 82.27 $\pm$ 0.42 & 6.86 $\pm$ 0.10 \\
\quad LambdaMART & 42.96 $\pm$ 0.27 & 64.74 $\pm$ 0.34 & 80.52 $\pm$ 0.22 & 7.52 $\pm$ 0.04 \\
\midrule
\multicolumn{5}{l}{\textbf{Spectraverse, PubChem} ($n=46{,}042$)} \\

FLARE & 18.74 & 44.55 & 68.76 & 6.93 \\
JESTR & 5.69 & 17.78 & 39.02 & 12.73 \\
MS-AGG & & & & \\
\quad Linear pairwise & 34.47 $\pm$ 1.86 & 61.56 $\pm$ 2.03 & 80.43 $\pm$ 0.71 & 4.93 $\pm$ 0.11 \\
\quad MLP$^{\dagger}$ & \textbf{41.73 $\pm$ 0.42} & \textbf{66.62 $\pm$ 0.18} & \textbf{82.66 $\pm$ 0.27} & \textbf{4.50 $\pm$ 0.02} \\
\quad YetiRank & 31.94 $\pm$ 0.24 & 58.60 $\pm$ 0.19 & 78.91 $\pm$ 0.25 & 5.20 $\pm$ 0.03 \\
\quad LambdaMART & 32.39 $\pm$ 0.15 & 60.02 $\pm$ 0.39 & 80.40 $\pm$ 0.16 & 5.13 $\pm$ 0.02 \\
\midrule
\multicolumn{5}{l}{\textbf{Spectraverse, CLM decoys} ($n=45{,}132$)} \\
FLARE & 13.78 & 30.77 & 47.96 & 8.08 \\
JESTR & 3.44 & 11.43 & 23.43 & 10.79 \\
MS-AGG & & & & \\
\quad Linear pairwise & 29.00 $\pm$ 0.02 & 49.99 $\pm$ 0.03 & 67.07 $\pm$ 0.05 & 5.57 $\pm$ 0.00 \\
\quad MLP$^{\dagger}$ & \textbf{39.88 $\pm$ 0.66} & \textbf{60.48 $\pm$ 1.03} & \textbf{74.46 $\pm$ 0.99} & \textbf{4.90 $\pm$ 0.08} \\
\quad YetiRank & 34.02 $\pm$ 0.09 & 55.40 $\pm$ 0.04 & 70.75 $\pm$ 0.09 & 5.24 $\pm$ 0.01 \\
\quad LambdaMART & 33.60 $\pm$ 0.33 & 56.05 $\pm$ 0.14 & 72.32 $\pm$ 0.13 & 5.23 $\pm$ 0.03 \\
\bottomrule
\end{tabular}

\begin{minipage}{0.98\linewidth}
\scriptsize Learned-ranker results are mean $\pm$ sample standard deviation across four seeds. $\dagger$ indicates the model selected by highest held-out validation Hit@1. Higher Hit@$k$ and lower MCES@1 indicate better performance. Bold entries indicate the best test value.

\end{minipage}\end{table}
\clearpage

\subsection*{MS-AGG Outperforms Aggregation Baselines}

We next compared MS-AGG against fixed-rule aggregation methods and a learned pointwise logistic regression baseline across the five candidate-pool settings (Table~\ref{tab:ablation}). Among the fixed rules, global z-mean achieved the best performance across nearly all settings and evaluation metrics, followed generally by within-pool z-mean and rank product. Rank product consistently outperformed reciprocal rank fusion and Borda count, with Borda count yielding the lowest performance. Global z-mean improved Hit@1 over within-pool z-mean by 0.70--2.96 percentage points across the five settings.

Pointwise logistic regression outperformed all fixed aggregation rules across all four metrics in four of the five candidate-pool settings, demonstrating the benefits of learning an aggregation function. However, it underperformed global z-mean across all four metrics on Spectraverse CLM decoys and failed to improve upon the strongest individual annotation tool in Hit@1 in two of the five settings.

All four MS-AGG rankers outperformed every aggregation baseline across all five candidate-pool settings and all four reported metrics. The highest-performing MS-AGG rankers improved Hit@1 over the strongest aggregation baseline by 6.95--16.70 percentage points across the five settings. Even the weakest MS-AGG ranker exceeded the strongest aggregation baseline by at least 0.58 percentage points in Hit@1 across the five settings. Consistent improvements were also observed in Hit@5, Hit@20, and MCES@1. These results highlight the benefits of explicitly optimizing relative candidate rankings during aggregation learning.

\begin{table}[H]
\caption{Fixed and learned aggregation across five candidate-pool settings.}
\label{tab:ablation}
\centering
{\color{black}\scriptsize
\setlength{\tabcolsep}{5pt}
\renewcommand{\arraystretch}{0.9}
\begin{tabular}{lrrrr}
\toprule
Method & Hit@1 (\%) $\uparrow$ & Hit@5 (\%) $\uparrow$ & Hit@20 (\%) $\uparrow$ & MCES@1 $\downarrow$ \\
\midrule
\multicolumn{5}{l}{\textbf{MassSpecGym, mass-based pools} ($n=17{,}556$)} \\
Global z-mean & 30.68 & 51.07 & 71.65 & 14.75 \\
Within-pool z-mean & 29.98 & 50.56 & 70.63 & 15.25 \\
Rank product\cite{gupta2026confronting} & 27.65 & 49.18 & 69.48 & 15.43 \\
RRF ($k=60$)\cite{Cormack2009RRF} & 24.19 & 45.14 & 66.11 & 15.81 \\
Borda count & 20.51 & 39.97 & 60.96 & 16.47 \\
Pointwise logistic & \underline{40.53} & \underline{67.12} & \underline{85.00} & \underline{13.08} \\
MS-AGG (worst--best) & 52.45--\textbf{54.66} & 74.88--\textbf{76.05} & 87.97--\textbf{89.09} & 9.93--\textbf{8.85} \\
\midrule
\multicolumn{5}{l}{\textbf{MassSpecGym, formula pools} ($n=17{,}556$)} \\
Global z-mean & 30.37 & 50.30 & 70.89 & 8.84 \\
Within-pool z-mean & 27.81 & 47.24 & 67.54 & 9.52 \\
Rank product & 26.20 & 46.72 & 66.62 & 9.93 \\
RRF ($k=60$) & 22.99 & 42.80 & 63.51 & 10.01 \\
Borda count & 20.97 & 39.34 & 60.13 & 10.10 \\
Pointwise logistic & \underline{40.38} & \underline{67.45} & \underline{85.25} & \underline{6.38} \\
MS-AGG (worst--best) & 46.25--\textbf{47.33} & 72.57--\textbf{73.07} & 88.21--\textbf{89.24} & 5.11--\textbf{4.93} \\
\midrule
\multicolumn{5}{l}{\textbf{MassSpecGym, CLM decoys} ($n=16{,}812$)} \\
Global z-mean & 26.05 & 45.05 & 63.36 & 10.33 \\
Within-pool z-mean & 25.05 & 44.25 & 63.38 & 10.54 \\
Rank product & 22.22 & 42.84 & 62.18 & 10.92 \\
RRF ($k=60$) & 19.29 & 38.59 & 58.95 & 11.24 \\
Borda count & 17.04 & 35.18 & 55.21 & 11.44 \\
Pointwise logistic & \underline{33.18} & \underline{58.14} & \underline{75.64} & \underline{9.23} \\
MS-AGG (worst--best) & 42.96--\textbf{49.88} & 64.74--\textbf{69.37} & 80.52--\textbf{82.31} & 7.52--\textbf{6.46} \\
\midrule
\multicolumn{5}{l}{\textbf{Spectraverse, PubChem} ($n=46{,}042$)} \\
Global z-mean & 29.44 & 54.90 & 73.39 & 5.87 \\
Within-pool z-mean & 26.48 & 52.64 & 71.61 & 6.32 \\
Rank product & 26.50 & 52.89 & 71.82 & 6.25 \\
RRF ($k=60$) & 25.06 & 51.33 & 70.87 & 6.30 \\
Borda count & 21.37 & 45.08 & 64.97 & 6.69 \\
Pointwise logistic & \underline{31.36} & \underline{57.68} & \underline{77.82} & \underline{5.45} \\
MS-AGG (worst--best) & 31.94--\textbf{41.73} & 58.60--\textbf{66.62} & 78.91--\textbf{82.66} & 5.20--\textbf{4.50} \\
\midrule
\multicolumn{5}{l}{\textbf{Spectraverse, CLM decoys} ($n=45{,}132$)} \\
Global z-mean & \underline{25.93} & \underline{47.94} & \underline{63.58} & \underline{6.17} \\
Within-pool z-mean & 24.01 & 46.09 & 61.93 & 6.49 \\
Rank product & 23.93 & 46.15 & 62.84 & 6.50 \\
RRF ($k=60$) & 23.05 & 44.30 & 61.12 & 6.57 \\
Borda count & 18.41 & 37.21 & 53.15 & 6.99 \\
Pointwise logistic & 22.98 & 42.41 & 59.63 & 6.35 \\
MS-AGG (worst--best) & 29.00--\textbf{39.88} & 49.99--\textbf{60.48} & 67.07--\textbf{74.46} & 5.57--\textbf{4.90} \\
\bottomrule
\end{tabular}
\begin{minipage}{0.98\linewidth}
\scriptsize
Each non-MS-AGG method combines all 19 MassSpecGym features or all 18 Spectraverse features.
MS-AGG ranges span the four MS-AGG ranker means in Table~\ref{tab:retrieval}.
\textbf{Bold} indicates the best MS-AGG performance, and \underline{underlining}
indicates the best non-MS-AGG performance for each metric and candidate-pool setting.
\end{minipage}}
\end{table}
\newpage

\subsection{Derived Features Add Predictive Information Beyond Annotation Scores}

To investigate the sources of improvement in MS-AGG, we systematically evaluated the predictive contributions of derived features beyond annotation scores, the effectiveness of different aggregation methods in exploiting these features, and the relative contributions of individual annotation tools to retrieval performance.

We investigated whether the derived features provide additional information that improves retrieval. We fit a range of models using only the three annotation scores, only the 16 derived features, and all 19 features on the MassSpecGym formula pools (Table~\ref{tab:matched-features}). 
When used independently, the annotation scores achieve higher Hit@1 than the derived features across all four aggregation methods. The best annotation-score-only result is $40.72 \pm 0.01\,\%$ Hit@1, compared with $24.91\,\%$ for the best derived-feature-only result.
Combining the two feature groups increases LambdaMART's Hit@1 to $46.98 \pm 0.21\,\%$, an improvement of 9.05 percentage points over its annotation scores alone. The molecule-clustered 95\% confidence interval for this improvement is $[6.23, 12.81]$ percentage points (Table~\ref{tab:stats}), providing evidence that the derived features contain information complementary to the annotation scores. The linear pairwise ranker shows a similar improvement, with Hit@1 increasing from 40.72\% using annotation scores alone to 46.25\% using all 19 features, a gain of 5.53 percentage points. This demonstrates that the additional information provided by the derived features can also be exploited by a linear ranking model, without requiring nonlinear aggregation.
This suggests that the relative effectiveness of linear and nonlinear aggregation depends on the available feature set: the linear pairwise ranker's stronger performance with only three annotation scores may reflect its ability to learn a simple, stable weighting of these signals, whereas LambdaMART's more flexible tree-based model may be less effective in this low-dimensional setting.

However, the presence of complementary information does not guarantee improved retrieval.
Both fixed z-mean aggregation methods perform worse when derived features are included: Hit@1 decreases from $36.30\,\%$ to $30.37\,\%$ for global z-mean and from $36.98\,\%$ to $27.81\,\%$ for within-pool z-mean.
Thus, while the derived features improve both linear and nonlinear learned rankers, equally weighted aggregation fails to exploit their additional information. These findings indicate that the benefit of derived features depends not only on their informativeness but also on how their contributions are learned and combined.

\begin{table}[!htbp]
\caption{Contribution of annotation scores and derived features to molecular retrieval on MassSpecGym formula pools, measured by Hit@1 (\%).}
\label{tab:matched-features}
\centering
\small
\begin{tabular}{rlll}
\toprule
Method & Annotation scores (3) & Derived features (16) & All 19 \\
\midrule
LambdaMART & 37.93 $\pm$ 1.71 & 24.05 $\pm$ 0.29 & \textbf{46.98 $\pm$ 0.21} \\
Linear Pairwise ranker & \textbf{40.72 $\pm$ 0.01} & 24.62 $\pm$ 0.03 & 46.25 $\pm$ 0.02 \\
Global z-mean & 36.30 & \textbf{24.91} & 30.37 \\
Within-pool z-mean & 36.98 & 23.80 & 27.81 \\
\bottomrule
\end{tabular}
\end{table}

Having established that the annotation scores and derived features provide complementary information, we next examined whether all three annotation tools contribute equally to the learned combination.
To examine how individual annotation tools contribute to learned aggregation, we evaluated LambdaMART with the 16 derived features together with FLARE, ICEBERG, and JESTR scores individually and in combination (Table~\ref{tab:tool-ablation}). ICEBERG provides the largest individual gain, increasing mean Hit@1 from $24.05 \pm 0.29\,\%$ with derived features alone to $44.60 \pm 0.38\,\%$. Adding FLARE alongside ICEBERG further increases Hit@1 to $47.03 \pm 0.14\,\%$, indicating that the two annotation scores provide complementary information. Including JESTR yields a similar Hit@1 of $46.98 \pm 0.21\,\%$, a difference of $-0.05$ percentage points (95\% CI $[-0.31, 0.21]$; Table~\ref{tab:stats}).
Adding JESTR produces only small changes in Hit@5 (72.96\% to 73.07\%), Hit@20 (88.07\% to 88.21\%), and MCES@1 (4.95 to 4.93). These results suggest that ICEBERG provides the strongest individual annotation signal in this setting, while FLARE contributes additional information that improves retrieval. 
\begin{table}[!htbp]
\caption{Contribution of individual annotation-tool scores and their combinations to molecular retrieval on MassSpecGym formula pools, measured by Hit@1, Hit@5, Hit@20 (\%), and MCES@1}
\label{tab:tool-ablation}
\centering
\small
\resizebox{\textwidth}{!}{%
\begin{tabular}{lrrrr}
\toprule
Annotation scores added & Hit@1 (\%) & Hit@5 (\%) & Hit@20 (\%) & MCES@1 \\
\midrule
None (derived features only) & 24.05 $\pm$ 0.29 & 39.97 $\pm$ 0.29 & 59.07 $\pm$ 0.07 & 9.59 $\pm$ 0.04 \\
\hspace{1em} $+$ FLARE & 32.17 $\pm$ 0.12 & 56.58 $\pm$ 0.06 & 78.36 $\pm$ 0.08 & 7.83 $\pm$ 0.02 \\
\hspace{1em} $+$ ICEBERG & 44.60 $\pm$ 0.38 & 67.15 $\pm$ 0.24 & 82.53 $\pm$ 0.11 & 5.41 $\pm$ 0.02 \\
\hspace{1em} $+$ JESTR & 24.44 $\pm$ 0.30 & 42.98 $\pm$ 0.29 & 63.90 $\pm$ 0.33 & 9.39 $\pm$ 0.02 \\
\hspace{1em} $+$ FLARE $+$ ICEBERG & \textbf{47.03 $\pm$ 0.14} & 72.96 $\pm$ 0.10 & 88.07 $\pm$ 0.22 & 4.95 $\pm$ 0.02 \\
\hspace{1em} $+$ FLARE $+$ JESTR & 32.33 $\pm$ 0.28 & 57.09 $\pm$ 0.18 & 79.04 $\pm$ 0.17 & 7.75 $\pm$ 0.02 \\
\hspace{1em} $+$ ICEBERG $+$ JESTR & 44.44 $\pm$ 0.56 & 67.98 $\pm$ 0.40 & 83.54 $\pm$ 0.54 & 5.42 $\pm$ 0.09 \\
\hspace{1em} $+$ ICEBERG $+$ FLARE $+$ JESTR & 46.98 $\pm$ 0.21 & \textbf{73.07 $\pm$ 0.19} & \textbf{88.21 $\pm$ 0.16} & \textbf{4.93 $\pm$ 0.03} \\
\bottomrule
\end{tabular}}
\begin{minipage}{0.98\linewidth}
\footnotesize Values are mean $\pm$ sample standard deviation across four independent runs. Bold marks the highest mean Hit rate or lowest mean MCES@1.
\end{minipage}
\end{table}

\begin{figure}[H]
\centering
\includegraphics[width=\linewidth]{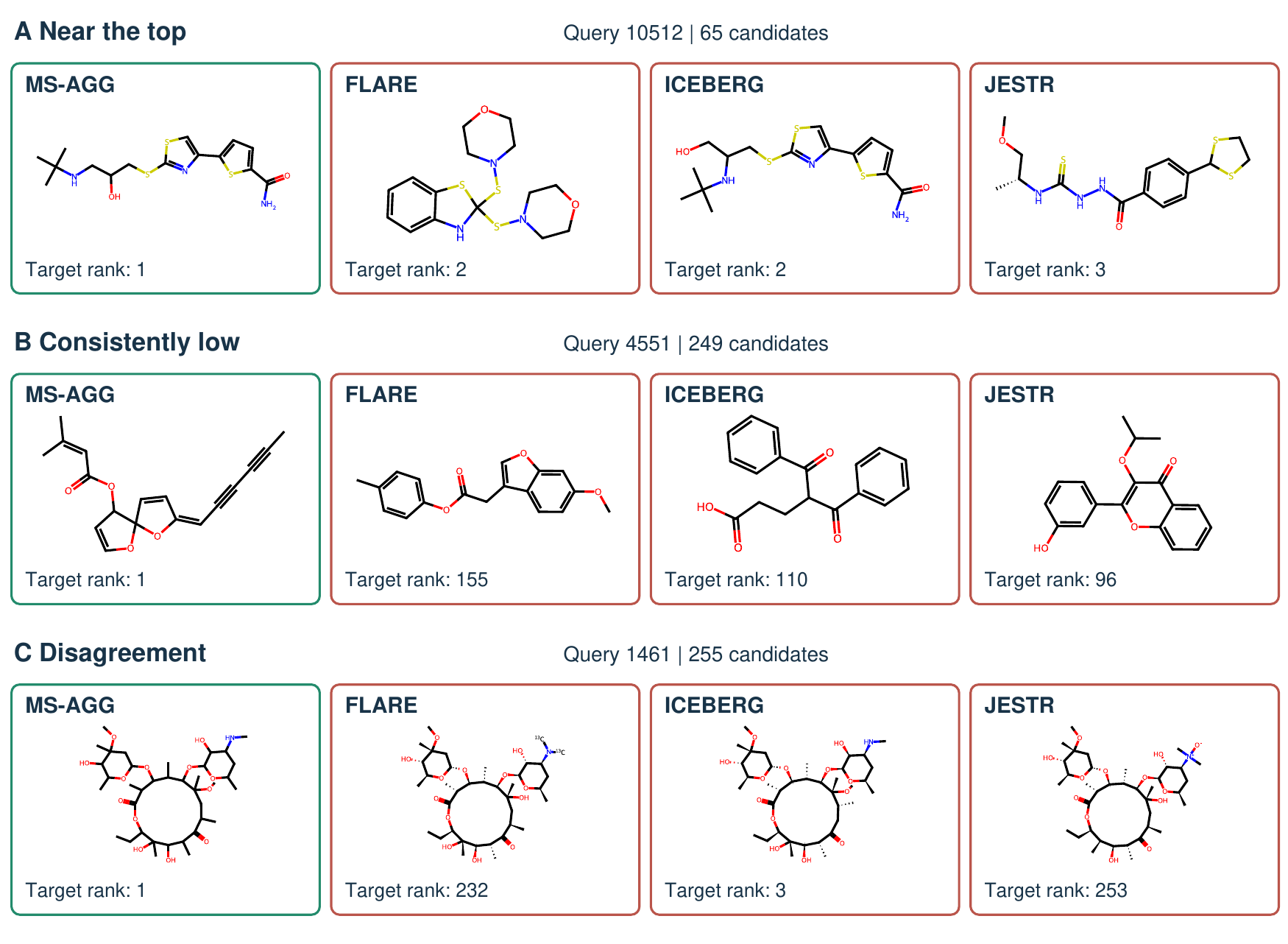}
\caption{Representative examples of molecular retrieval improved by learned aggregation over any individual feature's ranking. (A) the target ranks near the top for several features but is not ranked first by any; (B) the target consistently receives relatively low rankings across features; and (C) individual features assign substantially different ranks to the target. MS-AGG combines these complementary signals to recover the correct molecular structure. %
}
\label{fig:casestudy}
\end{figure}

\subsection{Learned Aggregation Recovers Missed Targets but Leaves Opportunities for Further Improvement}
\label{sec:aggregation-recovery}

Aggregation can recover target structures that no individual input feature ranks first. On MassSpecGym formula pools, the four-run LambdaMART score ensemble recovers 426 such queries ($2.4\,\%$ of test spectra), spanning 251 molecular identities. These cases show that combining candidate-level features and scores can identify the target even when every individual feature favors another candidate. Figure~\ref{fig:casestudy} provides illustrative examples of MS-AGG recovering the target when the annotation tools place the target near the top (A), consistently low (B), or at widely different ranks (C). In case (B), FLARE, ICEBERG and JESTR rank the target 155th, 110th, and 96th among 249 candidates, yet MS-AGG ranks it first.

Although learned aggregation improves overall molecular retrieval, these successes raise a broader question: how effectively does aggregation exploit the predictive information available across individual input features? To investigate this, we compared MS-AGG against a retrospective feature-selection \textit{oracle} that selects, for each query, the individual feature assigning the highest rank to the known target structure. This oracle represents the maximum performance achievable by selecting among the 19 individual feature rankings on a query-by-query basis, rather than a deployable retrieval method.

Across the 17,556 MassSpecGym formula-pool queries, both MS-AGG and the oracle correctly ranked the target first for 7,873 queries (44.8\%), while the oracle alone succeeded for 3,737 queries (21.3\%). Neither succeeded for 5,520 queries (31.4\%). Overall, the oracle achieved 66.1\% Hit@1, compared with 47.3\% for MS-AGG. 

These findings reveal both the potential and limitations of learned aggregation. While MS-AGG can recover targets missed by every individual feature, it does not consistently preserve the target predictions available from individual features. The gap relative to the retrospective oracle highlights opportunities to better exploit the information distributed across annotation scores and derived features, potentially through aggregation methods that adapt to the reliability and agreement of individual signals across candidate pools.

\section*{Conclusion}
Our work provides two contributions: (1) a ranking-based framework for learning to aggregate molecular annotation scores and features by explicitly optimizing relative candidate rankings; and (2) the development of derived features that capture predictive information from molecular fingerprints and spectrum--structure reference neighborhoods not explicitly represented by existing annotation tools. Across all candidate-pool settings on the MassSpecGym and Spectraverse benchmarks, all  MS-AGG rankers outperform the strongest individual annotation tool  and all aggregation baselines across all  evaluation metrics. Comparisons with pointwise logistic regression demonstrate the advantages of explicitly optimizing relative candidate rankings over learning an aggregation function through independent candidate predictions. Ablation studies further establish the value of the derived features and the importance of learning how to combine them with annotation scores.
Query-level analyses further demonstrate that MS-AGG can recover target structures missed by every individual feature, while the gap relative to a retrospective feature-selection oracle reveals opportunities to better exploit complementary predictive information. 
As agentic AI systems increasingly facilitate the execution of multiple scientific tools \cite{m2024augmenting,Chen2026MetaboAnnotate}, learned aggregation methods such as MS-AGG could provide specialized computational skills for integrating the predictions of complementary annotation tools. Our findings underscore the importance of learned aggregation in realizing the full potential of these emerging workflows.

As agentic AI systems increasingly facilitate the execution of multiple annotation tools \cite{Chen2026MetaboAnnotate}, learned aggregation methods such as MS-AGG could provide specialized computational skills for integrating their predictions. Our findings underscore the importance of learned aggregation in realizing the full potential of these emerging workflows.

Future work could move beyond the current set of hand-crafted derived features by learning feature representations directly from annotation scores, candidate molecular structures, and their relationships. Such representations could capture nonlinear interactions and patterns of agreement or disagreement among annotation tools that are not explicitly encoded in the current feature set. Another important direction is learning aggregation models that generalize across datasets. Our exploratory cross-dataset analyses suggest that the predictive utility of individual features varies across datasets, limiting the direct transferability of learned aggregation models across datasets. Future work could investigate transfer learning and domain adaptation strategies that capture shared predictive relationships while accommodating dataset-specific differences in feature effectiveness. Additional experimental information, including retention time, ion mobility, and spectra acquired under different fragmentation conditions, could further enrich these learned representations and improve molecular retrieval.

\section*{Data and Code Availability}

The MS-AGG code and reproducibility materials have been assembled for release (https://github.com/HassounLab/MS-AGG). They include training and evaluation code, model configurations, candidate-level features, predictions, and a manifest linking these files to the reported analyses. MassSpecGym and Spectraverse are available through their publications \cite{Bushuiev2024MSGym, gupta2026comprehensive}.

\section*{Supporting Information}

Section \ref{si:methods}  describes the implementation and selection of MS-AGG rankers and the implementation of fixed-rule and learned pointwise aggregation baselines. Section \ref{sec:suppData} describes the benchmark datasets, candidate pools. Section \ref{si:overlap} describes structural overlap between training and test sets. Section \ref{sec:suppResults} reports statistical comparisons, retrieval results by ionization mode and adduct, and analyses of annotation-score aggregation, and cross-dataset transfer.

\section*{Acknowledgments}

The authors thank the Tufts University High-Performance Computing cluster for computational resources and the developers of the public data sets and model implementations used in this work. Generative AI tools were used during manuscript preparation to suggest prose and formatting edits. The authors reviewed the resulting text and take responsibility for the experiments, numerical results, interpretations, and final manuscript.

\section*{Notes}

The authors declare no competing financial interest.

\section*{Funding}
Research reported in this publication was supported by the National Institute of General Medical Sciences of the National Institutes of Health under award number R35GM148219. The content is solely the responsibility of the authors and does not necessarily represent the official views of the NIH.

\bibliography{bibliography}

\newpage

\begin{center}
\noindent\textbf{For Table of Contents Only}\\[6pt]
\includegraphics[width=3.25in]{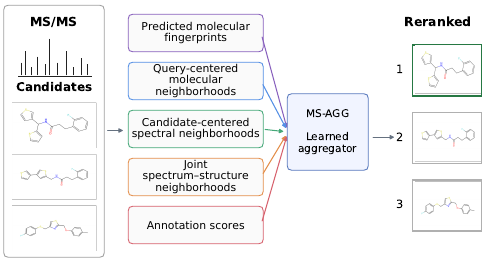}
\end{center}
\clearpage
\EnableSupplementSectionNumbers
\setcounter{page}{1}
\renewcommand{\thepage}{S-\arabic{page}}
\setcounter{section}{0}
\setcounter{table}{0}
\setcounter{figure}{0}
\setcounter{equation}{0}
\setcounter{secnumdepth}{3}
\renewcommand{\thesection}{S\arabic{section}}
\renewcommand{\thetable}{S\arabic{table}}
\renewcommand{\thefigure}{S\arabic{figure}}
\renewcommand{\theequation}{S\arabic{equation}}
\renewcommand{\theHsection}{supp.\arabic{section}}
\renewcommand{\theHtable}{supp.\arabic{table}}
\renewcommand{\theHfigure}{supp.\arabic{figure}}
\renewcommand{\theHequation}{supp.\arabic{equation}}
\begin{center}
{\Large\bfseries Supporting Information\par}
\medskip
{\large MS-AGG: Learning to Aggregate Tandem Mass Spectrometry Annotation Scores and Features for Improved Molecular Retrieval\par}
\medskip
Yinkai Wang, Yan Zhou Chen, and Soha Hassoun\par
Department of Computer Science, Tufts University, Medford, MA, USA\par
Department of Chemical and Biological Engineering, Tufts University, Medford, MA, USA\par
Corresponding author: \href{mailto:soha.hassoun@tufts.edu}{soha.hassoun@tufts.edu}
\end{center}
\section{Supplementary Methods}
\label{si:methods}
\subsection{Implementation of MS-AGG Rankers and Model Selection}
\label{si:ranker-methods}
\paragraph{Linear pairwise ranker.}
Features are standardized using validation-candidate means and standard deviations. The model uses scikit-learn logistic regression with L2 regularization ($C=1$), L-BFGS optimization, tolerance $10^{-4}$, and a maximum of 2{,}000 iterations.
The pairwise model learns to rank target-matching candidates above non-matching candidates by training on feature differences between the first target-matching candidate and up to 64 randomly sampled non-matching candidates from the same query. Each feature difference and its negative are assigned labels 1 and 0, respectively. The model is fitted without an intercept and ranks candidates using the learned linear score $w^{\mathsf T}x$.

\paragraph{MLP, LambdaMart, and YetiRank.}
The MLP has two 64-unit ReLU hidden layers and one scalar output, with listwise softmax cross-entropy, Adam at learning rate $10^{-3}$, five epochs, and one candidate group per update.  For MassSpecGym, the epoch with the highest held-out validation NDCG@10 is retained.

LambdaMART uses a learning rate of 0.03, a maximum of 31 leaves per tree, a maximum tree depth of 5, and a minimum of 200 samples per leaf. Feature subsampling (70\% of features per tree), data subsampling (80\% of samples every five boosting iterations), and L2 regularization (coefficient 1.0) are used to reduce overfitting. Training a 1,000-round LambdaMART model required approximately 1–2 minutes on eight CPU threads, excluding feature generation. Features were computed separately and cached for reuse across ranking models and experimental comparisons. 

YetiRank uses a maximum tree depth of 6 and a learning rate of 0.05. MassSpecGym tree rankers use NDCG@10 on held-out validation queries for checkpoint selection, with patience 30. Spectraverse YetiRank uses the same stopping criterion with a maximum of 200 iterations. 

Spectraverse candidate pools can exceed the group-size limits of the tree-based rankers. LightGBM's LambdaRank implementation accepts at most 10{,}000 rows per query group. The final Spectraverse PubChem validation pools contain as many as 42{,}792 candidates, with a 99th percentile of 11{,}015; evaluable test pools have a median of 338 candidates, a mean of 1{,}497, a maximum of 32{,}280, and a 99th percentile of 13{,}920. During training, groups above the LightGBM limit are reduced to 10{,}000 candidates. All positive candidates are retained, and the remaining positions are sampled uniformly from the negatives. CatBoost uses the same procedure with a configured cap of 5{,}000 rows per training query. Test pools are not subsampled. MassSpecGym is unaffected because its groups remain below both limits.

Within each benchmark validation split, 90\% of queries are used for fitting and 10\% for model selection. The split is made by query rather than molecular identity, so spectra from the same molecule may occur in both subsets.   For MassSpecGym, LambdaMART used up to 1,000 boosting rounds for formula and mass-based pools and up to 200 rounds for generated-decoy pools, while YetiRank used up to 1,000 rounds across all three settings. For Spectraverse, LambdaMART used 200 rounds and YetiRank used up to 200 rounds for both candidate-pool settings.

Among the MS-AGG four models, we select the one with the highest held-out Hit@1. Selected models are not refitted on the full validation set. For models trained with four seeds, test results report the mean and sample standard deviation across runs.

\subsection{Baseline Aggregation Models}
\label{si:fusion}

\paragraph{Pointwise logistic regression.}
The pointwise model uses the same scikit-learn logistic regression
configuration as the linear pairwise ranker. It learns to distinguish candidates matching the target structure from non-matching candidates, using balanced class weights and an intercept. At inference, candidates within each query are ranked by their predicted scores.

\paragraph{Rank-based aggregation.}

For a query $q$ with candidate set $\mathcal{C}_q$, let $s_f(c)$ denote the score assigned to candidate $c$ by feature $f$, and let $r_f(c)$ denote its rank in descending order of $s_f(c)$, with rank 1 being best. The set of features included in a comparison is denoted by $\mathcal{F}$.

We evaluate rank product, reciprocal rank fusion (RRF), and Borda count. Rank product orders candidates by the product of their feature-specific ranks, with smaller products ranked first. Equivalently, candidates can be ranked by the negative mean log rank:

\begin{equation}
g_{\mathrm{RP}}(c)
=
-\frac{1}{|\mathcal{F}|}
\sum_{f\in\mathcal{F}} \log r_f(c).
\end{equation}

RRF combines feature-specific ranks using a reciprocal-rank kernel with $k=60$:

\begin{equation}
g_{\mathrm{RRF}}(c)
=
\sum_{f\in\mathcal{F}}
\frac{1}{60+r_f(c)}.
\end{equation}

Borda count assigns each candidate points according to the number of candidates ranked below it:

\begin{equation}
g_{\mathrm{Borda}}(c)
=
\sum_{f\in\mathcal{F}}
\left(|\mathcal{C}_q|-r_f(c)\right).
\end{equation}

\paragraph{Z-score averaging.}
Z-score averaging combines standardized feature scores:

\begin{equation}
g_{\mathrm{z}}(c)
=
\frac{1}{|\mathcal{F}|}
\sum_{f\in\mathcal{F}}
\frac{s_f(c)-\mu_f}{\sigma_f}.
\end{equation}

We consider two normalization conventions. For within-pool z-score averaging, the mean and standard deviation are computed separately for each feature within each query's candidate pool. For global z-score averaging, one mean and standard deviation per feature are estimated across all validation candidates and applied to the test pools.

\section{Supplementary Dataset Analysis}
\label{sec:suppData}

\subsection{Benchmark Datasets and Candidate Sets}
Table~\ref{tab:datasets} summarizes the five benchmark settings, including the number of  test spectra used for evaluation, unique target  structures, and candidate-pool size distributions.

\begin{table}[H]
\caption{Data sets and candidate-pool statistics.}
\label{tab:datasets}
\centering
\footnotesize
\setlength{\tabcolsep}{3.4pt}%
\begin{tabular}{lrrr}
\toprule
Setting & Test spectra & Unique 2D structures & Pool size (med/mean/max) \\
\midrule
MassSpecGym, mass-based pools & 17{,}556 & 2{,}997 & 256/254/256 \\
MassSpecGym, formula pools & 17{,}556 & 2{,}997 & 218/161/256 \\
MassSpecGym, CLM decoys & 16{,}812 & 2{,}886 & 209/164/256 \\
Spectraverse, PubChem & 46{,}042 & 4{,}212 & 338/1{,}497/32{,}280 \\
Spectraverse, CLM decoys & 45{,}132 & 4{,}139 & 801/1{,}678/15{,}246 \\
\bottomrule
\end{tabular}
\begin{minipage}{0.98\linewidth}
\footnotesize Each benchmark uses the same test split across candidate constructions. Counts include spectra with the target structure in the pool; MassSpecGym generated-decoy pools additionally require at least one formula-matched decoy. Unique 2D structures are counted by the first InChIKey block of the target structure. Spectraverse pools are not capped at test time.
\end{minipage}
\end{table}

\noindent\textbf{MassSpecGym mass-based pools.} We used the original MassSpecGym mass-based candidate lists, with canonicalization and deduplication.\textsuperscript{\ref{supp-Bushuiev2024MSGym}} The MIST-CF predicted precursor formulas 
replaced the  formula in FLARE's precursor and peak-subformula inputs, the four predicted-fingerprint similarities, and the spectral-neighbor fingerprint feature. The remaining derived features and the frozen ICEBERG and JESTR scores were unchanged. FLARE's structure featurizer could not parse 19 validation and 31 test spectrum--candidate pairs. These pairs were excluded from FLARE inference but retained in the evaluation pools, with missing FLARE scores set one unit below the lowest valid score in each pool (or zero if none was available).
\medskip

\noindent\textbf{MassSpecGym formula-constrained pools.} The formula-constrained setting uses candidate structures consistent with the reference molecular formula. This setting evaluates molecular retrieval when the precursor formula is known, separating candidate ranking from precursor-formula prediction. The benchmark candidate pools are capped at 256 structures per spectrum.

\medskip

\noindent\textbf{MassSpecGym generated-decoy pools.} We trained a character-level recurrent-neural-network CLM\textsuperscript{\ref{supp-Skinnider2021CLM}} on MassSpecGym training-split structures and sampled new molecules. For each test query, we selected generated molecules matching the target formula, ordered them by sampling frequency, and supplemented them with released Spectraverse CLM candidates\textsuperscript{\ref{supp-gupta2026confronting}}. Candidates were deduplicated by 2D connectivity, excluding the target and its stereoisomers from the decoys. We retained up to 255 decoys, prioritizing our generated molecules, and added the target structure to form pools of at most 256 candidates. We excluded 744 spectra with no remaining decoy, leaving 16{,}812 test spectra.

\medskip

\noindent\textbf{Spectraverse PubChem pools.} We used the Spectraverse v1.0.0 spectral dataset\textsuperscript{\ref{supp-gupta2026comprehensive}}. Of its 46{,}159 test spectra, 46{,}042 retained the target structure and at least one alternative candidate after requiring the fingerprints needed for feature construction. The remaining 117 spectra were excluded.

\medskip

\noindent\textbf{Spectraverse generated-decoy pools.} 
We matched the 46{,}042 retained test queries to the released Spectraverse v1.0.2 generated-decoy pools using canonical target SMILES. \textsuperscript{\ref{supp-gupta2026confronting},\ref{supp-SpectraverseData102}} We added the target structure and removed duplicate canonical-SMILES strings. The 910 queries without any available non-target candidate were excluded, leaving 45{,}132 queries.

\section{Structural overlap between training and test sets}
\label{si:overlap}
Overlap between the training structures of constituent annotation models and benchmark test molecules could inflate retrieval performance and affect comparisons among models trained on different corpora. We therefore assessed exact structural overlap using canonical SMILES and near-overlap using Morgan fingerprints (radius 2, 4096 bits; Tanimoto similarity $\geq 0.95$).

For ICEBERG, we compared the 25,046 unique training structures from MassSpecGym against the 4,212 Spectraverse test reference structures, finding 36.56\% exact overlap and 53.23\% near-overlap. Given the substantial exact overlap, ICEBERG was excluded from Spectraverse evaluation.  In contrast, FLARE and JESTR used benchmark-specific training corpora with no exact training--test overlap; their shared Spectraverse training corpus exhibited 24.43\% near-overlap. 

Using 2D-connectivity identity instead of canonical SMILES identified six additional overlapping structures between Spectraverse training and test target structures, affecting 50 PubChem-pool and 38 decoy-pool spectra. Excluding these spectra did not change the ordering of the learned rankers.

\section{Supplementary Results}
\label{sec:suppResults}

\subsection{Statistical Significance of Retrieval Improvements}
\label{si:statistical-comparisons}

We assessed paired differences in Hit@1 between the validation-selected MS-AGG models and the strongest individual annotation scores and within-pool z-mean across five benchmark settings (Table~\ref{tab:stats-allsettings}). The within-pool z-mean was selected as a fixed-rule baseline because it combines standardized feature scores without learned weights, providing a direct comparison with learned aggregation. All comparisons used matched test queries and predictions from fitted models without retraining. Statistical significance was assessed using paired sign-flip tests with 10{,}000 random draws at the molecule level and a plus-one correction for $p$-values. The test evaluates whether the observed difference could arise under the null hypothesis of no systematic performance difference by randomly swapping the two methods' results for each target molecule and comparing the resulting differences with the observed difference.

Across the five settings, MS-AGG improved Hit@1 by 3.32--26.10
percentage points over the strongest individual annotation score
and by 15.26--22.48 percentage points over within-pool z-mean.
All 95\% confidence intervals excluded zero, indicating that
the improvements remained positive after accounting for
uncertainty due to the molecules included in the test set
(Table~\ref{tab:stats-allsettings}). The paired statistical
tests further supported these improvements, with $p=0.0022$
for MassSpecGym generated decoys versus ICEBERG and
$p<10^{-4}$ for all other comparisons.

\begin{table}[H]
\caption{Paired Hit@1 improvements of the validation-selected MS-AGG rankers over the strongest individual annotation score and within-pool z-mean across all five benchmark settings. }
\label{tab:stats-allsettings}
\centering
\small
\resizebox{\textwidth}{!}{%
\begin{tabular}{llrc}
\toprule
Setting & Comparison & $\Delta$ Hit@1 (pp) & 95\,\% CI \\
\midrule
\multirow{2}{*}{MassSpecGym, mass-based}
& MS-AGG $-$ ICEBERG & $+6.30$ & $[+4.20, +8.43]$ \\
& MS-AGG $-$ within-pool z-mean & $+22.48$ & $[+19.99, +24.99]$ \\
\midrule
\multirow{2}{*}{MassSpecGym, formula}
& MS-AGG $-$ ICEBERG & $+11.04$ & $[+7.77, +15.21]$ \\
& MS-AGG $-$ within-pool z-mean & $+19.17$ & $[+16.87, +21.48]$ \\
\midrule
\multirow{2}{*}{MassSpecGym, generated decoys}
& MS-AGG $-$ ICEBERG & $+3.32$ & $[+1.29, +5.48]$ \\
& MS-AGG $-$ within-pool z-mean & $+20.95$ & $[+18.27, +23.62]$ \\
\midrule
\multirow{2}{*}{Spectraverse, PubChem}
& MS-AGG $-$ FLARE & $+23.00$ & $[+20.80, +25.26]$ \\
& MS-AGG $-$ within-pool z-mean & $+15.26$ & $[+13.62, +16.86]$ \\
\midrule
\multirow{2}{*}{Spectraverse, CLM decoys}
& MS-AGG $-$ FLARE & $+26.10$ & $[+23.57, +28.60]$ \\
& MS-AGG $-$ within-pool z-mean & $+15.88$ & $[+14.31, +17.51]$ \\
\bottomrule
\end{tabular}
}
\begin{minipage}{0.98\textwidth}
\footnotesize
MS-AGG denotes LambdaMART for MassSpecGym mass- and formula-based pools, YetiRank for MassSpecGym decoys, and MLP for both Spectraverse settings. Differences use unrounded per-query correctness averaged across four seeds. Differences and molecule-clustered 95\% confidence intervals are in percentage points.
\end{minipage}
\end{table}

\subsection{Statistical Comparisons of Learned Models and Feature Configurations}
\label{si:model-feature-comparisons}

We examined whether differences among learned rankers and feature
configurations were statistically significant on the MassSpecGym
formula-constrained test set (Table~\ref{tab:stats}). Comparisons
included LambdaMART, YetiRank, and MLP, as well as LambdaMART
models trained using annotation scores alone, derived features
alone, or the full feature set. Statistical significance was
assessed using the paired procedures described in
Section~\ref{si:statistical-comparisons}.

Differences among LambdaMART, YetiRank, and MLP were not
statistically significant, suggesting that the choice among
these nonlinear rankers had limited influence on retrieval
performance in this setting. In contrast, combining annotation
scores and derived features significantly improved Hit@1
relative to using either feature group alone. The full
LambdaMART model improved Hit@1 by 9.05 percentage points
over the annotation-score-only model and by 22.94 points
over the derived-feature-only model, with both comparisons
yielding $p<10^{-4}$. These findings support the complementary
predictive value of annotation scores and derived features.
Additional comparisons with fixed-rule aggregation and a
reduced feature configuration are reported in
Table~\ref{tab:stats}.

\begin{table}[H]
\caption{Molecule-clustered paired Hit@1 comparisons among learned
rankers, fixed-rule aggregation baselines, and feature configurations
on the MassSpecGym formula-constrained test set. Positive differences
indicate higher Hit@1 for the first method in each comparison.}
\label{tab:stats}
\centering
\small
\resizebox{\textwidth}{!}{%
\begin{tabular}{lrcc}
\toprule
Comparison & $\Delta$ Hit@1 (pp) & 95\,\% CI & Permutation $p$ \\
\midrule
LambdaMART $-$ MLP
& $+0.67$ & $[-1.41, +3.75]$ & 0.871 \\
LambdaMART $-$ YetiRank
& $-0.35$ & $[-1.12, +0.39]$ & 0.383 \\
YetiRank $-$ MLP
& $+1.02$ & $[-1.21, +4.37]$ & 0.770 \\
\midrule
LambdaMART $-$ global z-mean
& $+16.61$ & $[+14.53, +18.71]$ & $<10^{-4}$ \\
\midrule
Full $-$ annotation-score-only LambdaMART
& $+9.05$ & $[+6.23, +12.81]$ & $<10^{-4}$ \\
Full $-$ derived-feature-only LambdaMART
& $+22.94$ & $[+20.67, +25.38]$ & $<10^{-4}$ \\
Derived-feature-only LambdaMART $-$ global z-mean
& $-6.33$ & $[-7.85, -4.83]$ & $<10^{-4}$ \\
Full $-$ (derived + FLARE + ICEBERG)
& $-0.05$ & $[-0.31, +0.21]$ & 0.720 \\
\bottomrule
\end{tabular}
}
\begin{minipage}{0.98\textwidth}
\footnotesize
The 17{,}556 test spectra correspond to 2{,}997 target
molecules. Paired Hit@1 differences were computed from
predictions on the same test spectra, averaging across four
random seeds. The 95\% confidence intervals were estimated
using molecule-clustered bootstrap resampling, retaining all
spectra associated with each sampled target molecule.
Permutation $p$-values were obtained using paired sign-flip
tests at the molecule level.
\end{minipage}
\end{table}

\subsection{Retrieval Across Ionization Modes and Adducts}

To examine whether the improvements from learned aggregation
extend across ionization conditions, we stratified the
Spectraverse v1.0.0 test spectra with PubChem candidate pools
by ionization mode and precursor adduct. Table~\ref{tab:sv-adduct}
reports Hit@1 for the validation-selected MS-AGG MLP,
individual FLARE and JESTR input scores, and two fixed
aggregation rules: rank product and within-pool z-mean.

MS-AGG achieved 39.2\% Hit@1 in positive mode and 47.6\% in negative mode, outperforming all comparison methods in both settings.
Relative to within-pool z-mean, MS-AGG improved Hit@1 by
14.94 percentage points in positive mode (molecule-clustered
95\% CI: 13.25--16.76) and 15.99 points in negative mode
(13.73--18.22). In contrast, individual annotation scores
showed substantial differences between modes, particularly
FLARE, whose Hit@1 decreased from 20.7\% in positive mode
to 14.2\% in negative mode. These findings suggest that
learned aggregation can mitigate limitations of individual
annotation scores by combining information from multiple
features.

Performance also varied across precursor adducts.
MS-AGG achieved the highest Hit@1 among the compared
methods in all eight reported adduct categories, although
its advantage differed across strata. For example, MS-AGG
achieved 42.6\% Hit@1 for protonated molecules
($[\mathrm{M}+\mathrm{H}]^{+}$), compared with 27.3\%
for sodium adducts ($[\mathrm{M}+\mathrm{Na}]^{+}$).
The smaller sample sizes for some adduct categories
warrant caution when interpreting these differences.
Nevertheless, the results indicate that the benefits
of learned aggregation extend beyond the predominant
protonated and deprotonated ions.

\begin{table}[H]
\centering
\caption{Spectraverse Hit@1 by ionization mode and adduct.}
\label{tab:sv-adduct}
\small
\resizebox{\textwidth}{!}{%
\begin{tabular}{lrrrrrr}
\toprule
Stratum & $n$ & MS-AGG & FLARE input & JESTR input & Rank product & z-mean (18) \\
\midrule
All & 46{,}042 & \textbf{41.7} & 18.7 & 5.2 & 13.4 & 26.5 \\
\midrule
Positive mode & 32{,}123 & \textbf{39.2} & 20.7 & 5.1 & 14.1 & 24.3 \\
Negative mode & 13{,}919 & \textbf{47.6} & 14.2 & 5.4 & 11.9 & 31.6 \\
\midrule
$[\mathrm{M}+\mathrm{H}]^{+}$                 & 21{,}463 & \textbf{42.6} & 24.6 & 5.3 & 16.5 & 25.2 \\
$[\mathrm{M}-\mathrm{H}]^{-}$                 & 10{,}765 & \textbf{47.5} & 16.9 & 5.6 & 13.7 & 32.1 \\
$[\mathrm{M}+\mathrm{Na}]^{+}$                &  6{,}930 & \textbf{27.3} & 16.6 & 5.1 & 11.1 & 16.5 \\
$[\mathrm{M}+\mathrm{NH}_{4}]^{+}$            &  3{,}029 & \textbf{45.3} &  4.8 & 3.4 &  4.6 & 36.7 \\
$[\mathrm{M}+\mathrm{FA}-\mathrm{H}]^{-}$     &  1{,}625 & \textbf{51.2} &  5.8 & 6.9 &  8.4 & 26.3 \\
$[\mathrm{M}+\mathrm{CH}_{3}\mathrm{COO}]^{-}$ &      825 & \textbf{47.6} &  2.3 & 1.5 &  0.8 & 44.5 \\
$[\mathrm{M}+\mathrm{Cl}]^{-}$                &      704 & \textbf{41.0} &  6.7 & 3.8 &  5.4 & 20.9 \\
$[\mathrm{M}+\mathrm{K}]^{+}$                 &      426 & \textbf{16.8} &  8.5 & 4.0 &  7.3 & 16.0 \\
\bottomrule
\end{tabular}%
}
\smallskip
\begin{minipage}{0.98\textwidth}
    \footnotesize
    Hit@1 (\%) uses 2D-connectivity identity. MS-AGG is the four-seed
    mean for the validation-selected five-epoch MLP. The other columns
    report frozen input-score or fixed-rule rankings. Rank product
    combines the FLARE and JESTR input ranks. Within-pool z-mean
    combines all 18 features. The JESTR input score differs from the
    published JESTR result in Table~2 of the main manuscript. Bold
    marks the largest value in each stratum. $n$ is the number of spectra.
\end{minipage}
\end{table}

\subsection{Cross-Dataset Feature and Transfer Analyses}
\label{si:transfer-results}

To investigate the transferability of MS-AGG across datasets, we examined whether individual features retain their predictive utility across MassSpecGym and Spectraverse, whether learned aggregation models transfer between the two benchmarks, and whether fine-tuning or joint training improves cross-dataset performance. These exploratory analyses assess the extent to which learned aggregation depends on dataset-specific characteristics.

\noindent\textbf{Cross-dataset feature effectiveness.}
We compared the individual Hit@1 performance of the 18 features shared by MassSpecGym and Spectraverse. The Spearman correlation between feature rankings was modest ($\rho=0.36$), indicating that their relative effectiveness differed across datasets. For example, \texttt{pn\_msn\_spec} achieved similar Hit@1 values on MassSpecGym and Spectraverse (15.9\% and 17.0\%, respectively), whereas \texttt{s\_ap} achieved 6.8\% and 17.3\%, and \texttt{nn\_max\_K100} achieved 20.4\% and 13.2\%. These differences suggest that the predictive utility of individual features is dataset-dependent, potentially limiting the direct transfer of learned aggregation models.

\noindent\textbf{Direct transfer and fine-tuning.}
We evaluated transfer in both directions using single-seed LambdaMART models trained for 200 rounds with the 18 shared features. These experiments evaluated performance on 17{,}556 MassSpecGym test spectra and 46{,}042 Spectraverse spectra, using canonical-SMILES matching. A model trained on Spectraverse achieved 26.01\% Hit@1 on MassSpecGym, compared with 32.32\% for a model trained on MassSpecGym. Fine-tuning the transferred model for an additional 100 rounds on MassSpecGym validation queries increased Hit@1 to 31.17\%, recovering much of the performance difference. In the reverse direction, a MassSpecGym-trained model achieved 22.05\% Hit@1 on Spectraverse, compared with 31.56\% for a model trained on Spectraverse. Fine-tuning in this direction was not evaluated. Thus, direct transfer reduced retrieval performance in both directions, while target-dataset fine-tuning substantially improved performance in the direction examined.

\noindent\textbf{Joint training.}
We also evaluated whether training on both datasets improved retrieval relative to training on each dataset separately. Joint training achieved 29.35\% Hit@1 on MassSpecGym and 31.60\% on the Spectraverse, compared with 32.32\% and 31.56\%, respectively, for the corresponding in-domain models. Joint training therefore reduced Hit@1 by 2.97 percentage points on MassSpecGym and increased it by only 0.04 points on Spectraverse, providing no consistent improvement.

These results suggest that differences in feature effectiveness across datasets may limit the transferability of learned aggregation models. Fine-tuning on target-dataset examples recovered much of the performance lost during direct transfer, whereas joint training provided no consistent benefit. Because these experiments used a single seed and a restricted feature set, the findings should be interpreted as exploratory rather than definitive conclusions about cross-dataset generalization.

\section*{References}
\begingroup
\renewcommand{\theenumi}{S\arabic{enumi}}
\renewcommand{\labelenumi}{(\theenumi)}
\begin{enumerate}
\item\label{supp-Bushuiev2024MSGym} Bushuiev, R.; Bushuiev, A.; de Jonge, N. F.; Young, A.; Kretschmer, F.; Samusevich, R.; Heirman, J.; Wang, F.; Zhang, L.; D{\"u}hrkop, K.; others. MassSpecGym: A Benchmark for the Discovery and Identification of Molecules. \emph{Advances in Neural Information Processing Systems} \textbf{2024}, \emph{37}, 110010--110027.
\item\label{supp-Skinnider2021CLM} Skinnider, M. A.; Stacey, R. G.; Wishart, D. S.; Foster, L. J. Chemical language models enable navigation in sparsely populated chemical space. \emph{Nature Machine Intelligence} \textbf{2021}, \emph{3}, 759--770.
\item\label{supp-gupta2026confronting} Gupta, V.; Xu, C.; Herbst, E.; Wang, F.; Wishart, D. S.; Skinnider, M. A. Confronting spurious evaluations of computational methods in small molecule mass spectrometry. \emph{bioRxiv} \textbf{2026}, Preprint, version 2, 14 July 2026.
\item\label{supp-gupta2026comprehensive} Gupta, V.; Qiang, H.; Chung, H.-H.; Herbst, E.; Skinnider, M. A. Comprehensive Curation and Harmonization of Small-Molecule MS/MS Libraries in Spectraverse. \emph{Analytical Chemistry} \textbf{2026}, \emph{98}, 3934--3943.
\item\label{supp-SpectraverseData102} Gupta, V.; Qiang, H.; Chung, H.-H.; Herbst, E.; Skinnider, M. Comprehensive Curation and Harmonization of Small Molecule MS/MS Libraries in Spectraverse. Zenodo, 2026; Data set, version v1.0.2.
\end{enumerate}
\endgroup

\end{document}